\documentclass{article}
\usepackage{spconf,amsmath,graphicx}
\usepackage{amssymb, amsmath}
\usepackage{algorithmic}
\usepackage{algorithm}
\usepackage{mathtools}
\usepackage{stmaryrd}
\usepackage{color}
\usepackage{multirow}
\usepackage{subcaption}

\usepackage{lipsum}

\newcommand\blfootnote[1]{%
  \begingroup
  \renewcommand\thefootnote{}\footnote{#1}%
  \addtocounter{footnote}{-1}%
  \endgroup
}

\newcommand{\ten}[1]{\boldsymbol{\mathcal{#1}}}

\title{{ADMM-MM Algorithm for General Tensor Decomposition}}
\name{Manabu Mukai$^1$, Hidekata Hontani$^1$, and Tatsuya Yokota$^{1,2}$}
\address{
$^1$ Nagoya Institute of Technology, Aichi, Japan\\
$^2$ RIKEN Center for Advanced Intelligence Project, Tokyo, Japan
}

\begin{document}
%
\maketitle
\begin{abstract}
In this paper, we propose a new unified optimization algorithm for general tensor decomposition which is formulated as an inverse problem for low-rank tensors in the general linear observation models.
The proposed algorithm supports three basic loss functions ($\ell_2$-loss, $\ell_1$-loss and KL divergence) and various low-rank tensor decomposition models (CP, Tucker, TT, and TR decompositions).
We derive the optimization algorithm based on hierarchical combination of the alternating direction method of multiplier (ADMM) and majorization-minimization (MM).
We show that wide-range applications can be solved by the proposed algorithm, and can be easily extended to any established tensor decomposition models in a {plug-and-play} manner.
\end{abstract}

\blfootnote{This work was supported in part by the Japan Society for the Promotion of Science (JSPS) KAKENHI under Grant 23H03419 and Grant 20H04249.}

\section{Introduction}
Tensor decompositions (TDs) are beginning to be used in various fields such as image recovery, blind source separation, traffic data analysis, and wireless communications \cite{yokota2021low,cichocki2007nmf,li2013efficient,chen2021tensor}.
Least-squares (LS) based algorithms have been well established for various kinds of TDs such as CP, Tucker, TT, TR, and non-negative tensor factorization \cite{kolda2009tensor,de2000best,holtz2012alternating,zhao2016tensor,cichocki2007nonnegative}.

\begin{figure}[t]
\centering
\includegraphics[width = 0.4\textwidth]{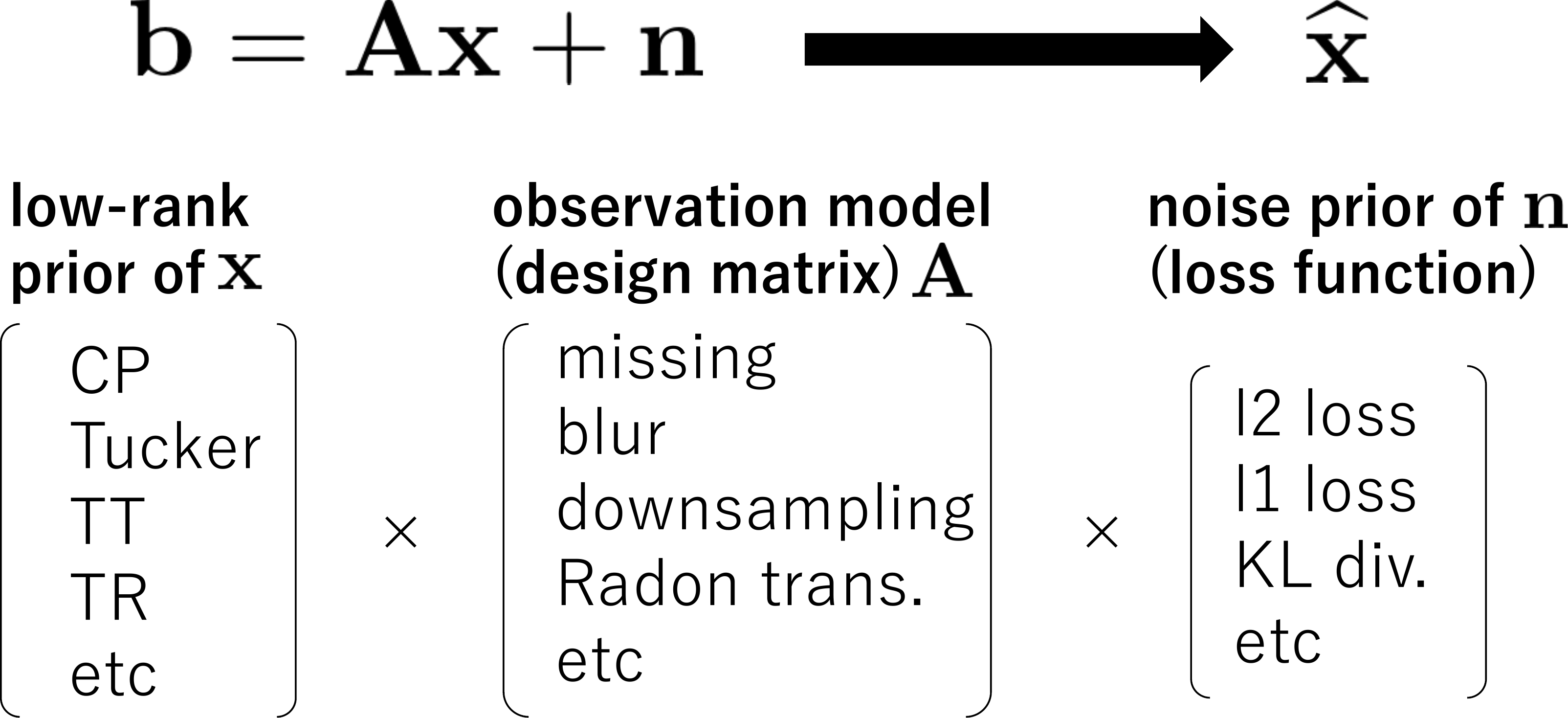}
\caption{General tensor decomposition problem}\label{fig:concept}
\end{figure}

There are two directions of generalization of TD problems.
One is to extend it for other specific application such as tensor completion, deblurring, super-resolution, computed tomography, and compressed sensing \cite{yokota2021low,shepp1982maximum,candes2005signal}.
The another is applying different loss functions for robust to different noise distributions \cite{shepp1982maximum,candes2011robust}.
Most of these generalizations are equivalent to solving an inverse problem with linear observation model:
\begin{align}
    \mathbf b = \mathbf A \mathbf x + \mathbf n,
    \label{eq:linear_obs_model}
\end{align}
where $\mathbf b \in \mathbb{R}^I$ is an observed signal, $\mathbf A \in \mathbb{R}^{I\times J}$ is a design matrix, $\mathbf x \in \mathbb{R}^J$ is unknown signal which has low-rank TD, and $\mathbf n \in \mathbb{R}^I$ is a noise component.
Introducing a low-rank TD constraint $\mathbf x \in \boldsymbol{\mathcal{S}}$ and noise assumption for $\mathbf n$, the optimization problem can be given as
\begin{align}
\mathop{\text{minimize}}_{\mathbf{x}}{D(\mathbf{b},\mathbf{A}\mathbf{x})}, \rm{s.t. } \ \mathbf {x} \in \boldsymbol{\mathcal{S}}, \label{eq:general_tensor_reconstruction_problem}
\end{align}
where $D(\cdot,\cdot)$ stands for loss function such as $\ell_2$-loss, $\ell_1$-loss and Kullback–Leibler (KL) divergence.
This is the general tensor decomposition (GTD) problem we aim to solve in this study (Fig.\ref{fig:concept}). 
Note that $\mathbf x \in \mathbb{R}^J$ is a $J$-dimensional vector in the form, but we consider $\mathbf x$ represents a tensor $\ten{X} \in \mathbb{R}^{J_1 \times J_2 \times \cdots \times J_N}$ by $\mathbf x = \text{vec}(\ten{X})$, where $J = \prod_{i=1}^N J_i$.

Solution algorithm for \eqref{eq:general_tensor_reconstruction_problem} has been studied only in limited cases.
For examples, research on tensor completion \cite{xu2015parallel,yokota2016smooth} is a case of specific $\mathbf A$ with $\ell_2$-loss, and robust TD \cite{zhang2013robust} is a case of specific $\mathbf A = \mathbf I$ with $\ell_1$-loss.
Several general optimization methods such as block coordinate descent (BCD) and projected gradient (PG) could be applied, but these were not designed to be well suited for \eqref{eq:general_tensor_reconstruction_problem}.

In this study, we propose a new algorithm for solving the GTD problem \eqref{eq:general_tensor_reconstruction_problem}.
The proposed algorithm supports three basic loss functions based on $\ell_2$-loss, $\ell_1$-loss and KL divergence with arbitrary design matrix $\mathbf A$, and can be applied to any TD models of which a least squares solution has been well established.
The proposed algorithm is derived from a hierarchical combination of alternating direction method of multiplier (ADMM) \cite{boyd2011distributed} and majorization-minimization (MM) framework \cite{hunter2004tutorial,sun2016majorization}.

Strength of our framework is its extensibility.
Even a completely new TD can be easily generalized to various design matrices and loss functions in a plug-and-play manner as long as the least-squares solution can be derived.

\section{APPROACH}

\subsection{Preliminary and Motivation}
Tensor decomposition (TD) is a mathematical model to represent a tensor  as a product of tensors/matrices.
For example, rank-$R$ CP decomposition of tensor $\boldsymbol{\mathcal{X}} \in \mathbb{R}^{J_1 \times J_2 \times J_3}$ is given by $\boldsymbol{\mathcal{X}}(\mathbf U_1, \mathbf U_2, \mathbf U_3) = \llbracket \mathbf{U}_1,\mathbf{U}_2,\mathbf{U}_3\rrbracket$,
where $\mathbf U_1 \in \mathbb{R}^{J_1 \times R}$, $\mathbf U_2 \in \mathbb{R}^{J_2 \times R}$, and $\mathbf U_3 \in \mathbb{R}^{J_3 \times R}$ are factor matrices.
More generally, let the set of factor tensors/matrices be $\theta \in \Theta$, and write the tensor by $\boldsymbol{\mathcal{X}}(\theta)$.

The most basic way for approximating a tensor $\boldsymbol{\mathcal{V}}$ with a TD model is the least squares (LS) method: 
\begin{align}
\hat{\theta} = \mathop{\text{argmin}}_{\theta \in \Theta} || \text{vec}(\boldsymbol{\mathcal{V}}) - \text{vec}(\boldsymbol{\mathcal{X}}(\theta)) ||_2^2.
\end{align}
Let be $\mathbf v = \text{vec}(\boldsymbol{\mathcal{V}})$, and the range of (vectorized form of) tensor be $\boldsymbol{\mathcal{S}} := \{ \text{vec}(\boldsymbol{\mathcal{X}}(\theta) ) \ | \ \theta \in \Theta \}$, we have
\begin{align}
  \text{vec}(\boldsymbol{\mathcal{X}}(\hat{\theta})) = \mathop{\text{argmin}}_{\mathbf x \in \boldsymbol{\mathcal{S}}} || \mathbf v - \mathbf x ||_2^2 =: \text{proj}_{\boldsymbol{\mathcal{S}}}(\mathbf v). \label{eq:LSTD}
\end{align}
Thus, LS-based TD can be regarded as a projection of $\mathbf v$ onto the set of low-rank tensors $\boldsymbol{\mathcal{S}}$.
In this study, we assume that $\text{proj}_{\boldsymbol{\mathcal{S}}}(\cdot)$ exists or can be replaced by some established TD algorithm such as alternating least squares (ALS) \cite{kolda2009tensor}.

The aim of this study is to solve the GTD problem \eqref{eq:general_tensor_reconstruction_problem} which is a generalization of LS-based TD \eqref{eq:LSTD}.
The GTD problem includes various TD models, various design matrices, and various loss functions, and its number of combinations is enormous.
Although there are many existing studies that derive optimization algorithms individually for special cases, the purpose of this study is to develop a single solution algorithm for the entire problems, including existing and unsolved problems.

\subsection{Sketch of Optimization Framework}
\begin{figure}[t]
\centering
\includegraphics[width=0.45\textwidth]{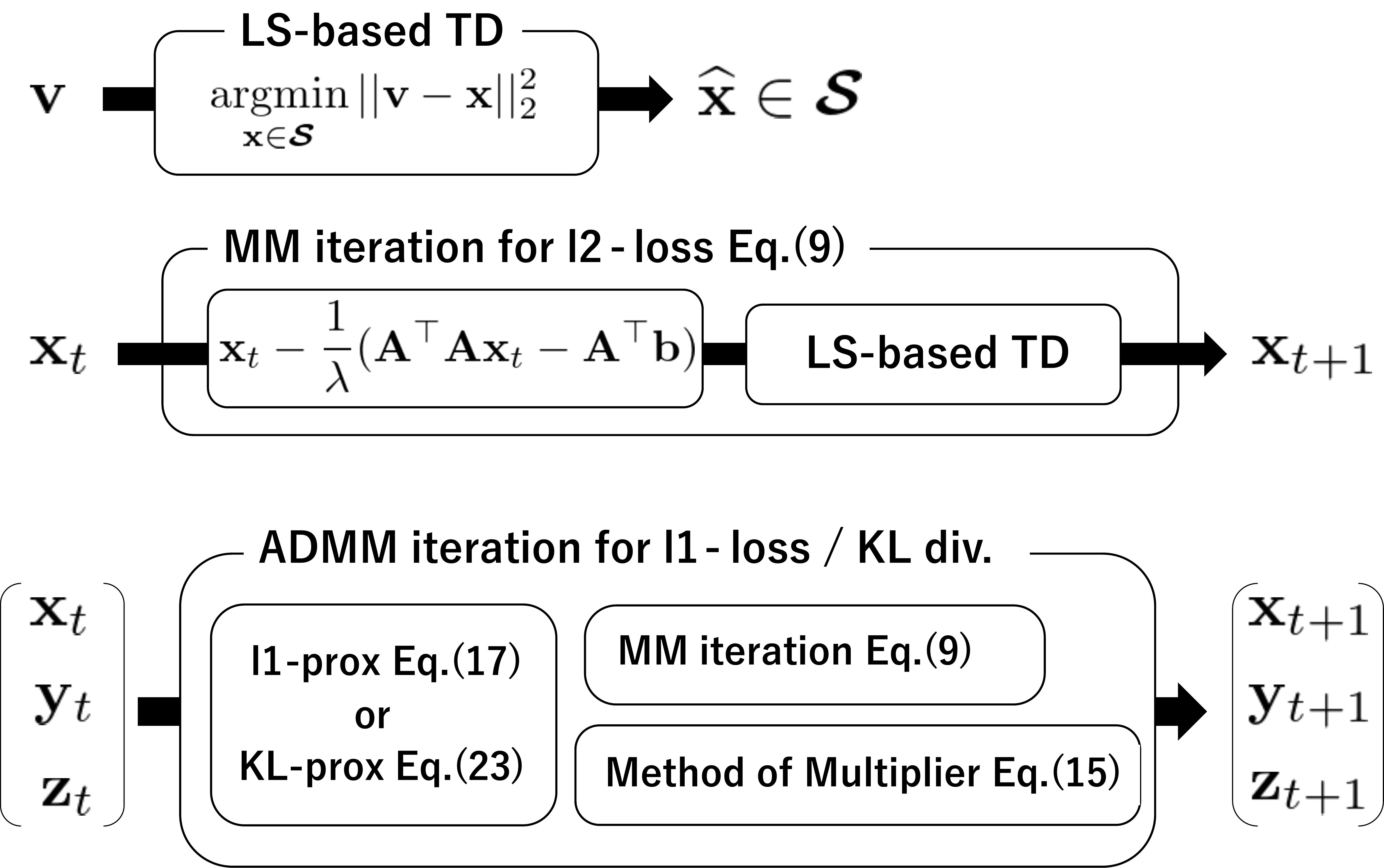}
\caption{Sketch of Optimization Algorithm}\label{fig:sketch}
\end{figure}

The key idea for solving a diverse set of problems by a simple algorithm is to use an already solvable problem to solve another extended problem. 
Fig.~\ref{fig:sketch} shows the sketch of optimization framework.
In this study, we first solve the case of $\ell_2$-loss using LS-based TD with MM framework.
Furthermore, we use it to solve the cases of $\ell_1$-loss and KL divergence with ADMM framework.
By replacing the module of LS-based TD, various types of TD can be easily generalized and applied to various applications.

\subsection{Optimization}
\subsubsection{MM for $\ell_2$-loss}\label{sec:L2case}
MM is an optimization framework for the objective function $f(\mathbf x)$ by iterative minimization of the auxiliary function: 
\begin{align}
  \mathbf x_{t+1} = \mathop{\text{argmin}}_{\mathbf x} g(\mathbf x|\mathbf x_t). \label{eq:MMiteration}
\end{align}
When the auxiliary function $g(\mathbf x|\mathbf x')$ satisfies $g(\mathbf x | \mathbf x') \geq f(\mathbf x)$ and $g(\mathbf x | \mathbf x)=f(\mathbf x)$, the MM iteration decreases the objective function $f(\mathbf x_{t+1}) \leq f(\mathbf x_t)$.

For the case of $\ell_2$-loss in \eqref{eq:general_tensor_reconstruction_problem}, objective function and its auxiliary function can be given by
\begin{align}
f(\mathbf x)            :=& || \mathbf b - \mathbf A \mathbf x ||_2^2 + i_{\boldsymbol{\mathcal{S}}}(\mathbf{x}), \label{eq:MMobj} \\
g(\mathbf x|\mathbf x') :=& f(\mathbf x) + (\mathbf x - \mathbf x')^\top (\lambda \mathbf I - \mathbf A^\top \mathbf A) (\mathbf x - \mathbf x') \label{eq:aux_func}\\
                        =& \lambda \left\|\mathbf x - \left(\mathbf{x}' - \frac{1}{\lambda} (\mathbf{A}^\top\mathbf{A}\mathbf{x}' - \mathbf{A}^\top\mathbf{b}) \right) \right\|^2 \notag \\ 
                        & + i_{\boldsymbol{\mathcal{S}}}(\mathbf{x}) + \rm{const},
\end{align}
where $i_{\boldsymbol{\mathcal{S}}}(\cdot)$ is an indicator function:
\begin{align}
i_{\boldsymbol{\mathcal{S}}}(\mathbf{x}) 
= \begin{cases}
    0 & \mathbf x \in \boldsymbol{\mathcal{S}} \\
    \infty & \mathbf x \notin \boldsymbol{\mathcal{S}} 
    \end{cases}.
\end{align}
From \eqref{eq:aux_func}, it is necessary that the additional term to be non-negative for $g(\mathbf x | \mathbf x') \geq f(\mathbf x)$.
It can be satisfied by setting $\lambda$ is larger than the maximum eigenvalue of $\mathbf A^\top \mathbf A$.
Then, the MM step \eqref{eq:MMiteration} can be reduced to 
\begin{align}
  \mathbf x_{t+1} = \text{proj}_{\boldsymbol{\mathcal{S}}}\left[ \mathbf{x}_t - \frac{1}{\lambda}(\mathbf{A}^\top\mathbf{A}\mathbf{x}_t - \mathbf{A}^\top \mathbf{b}) \right]. \label{eq:L2MMupdate}
\end{align}
Note that it is equivalent form to the PG algorithm with specific step-size $\frac{1}{\lambda}$.

Although it is written that $\mathbf x$ is updated in the formula, it is important to note that the TD parameter $\theta$ is actually updated by the algorithm in practice.
A flow including $\theta$ is given by
\begin{align}
 &\mathbf x_t \leftarrow \text{vec}( \ten{X}(\theta_t)), \\
 &\mathbf v_t \leftarrow \mathbf{x}_t - \frac{1}{\lambda}(\mathbf{A}^\top\mathbf{A}\mathbf{x}_t - \mathbf{A}^\top \mathbf{b}),\\
 &\theta_{t+1} \leftarrow \mathop{\text{argmin}}_{\theta \in \Theta} || 
 \mathbf v_t - \text{vec}(\boldsymbol{\mathcal{X}}(\theta)) ||_2^2,\\
 &\mathbf x_{t+1} \leftarrow \text{vec}( \ten{X}(\theta_{t+1})).
\end{align}

\subsubsection{ADMM for $\ell_1$-loss}\label{sec:L1case}
Employing $\ell_1$-loss for \eqref{eq:general_tensor_reconstruction_problem}, the problem is given as
\begin{align}
\mathop{\text{minimize}}_{\mathbf x} || \mathbf b - \mathbf A \mathbf x ||_1 + i_{\boldsymbol{\mathcal{S}}}(\mathbf x).
\end{align}
It can be transformed into the following equivalent problem:
\begin{align}
\mathop{\text{minimize}}_{\mathbf x, \mathbf y} || \mathbf b - \mathbf y ||_1 + i_{\boldsymbol{\mathcal{S}}}(\mathbf x), \text{s.t. } \mathbf y = \mathbf A \mathbf x.
\end{align}
Its augmented Lagrangian is given by
\begin{align}
L_{\rm L1}(\mathbf x,\mathbf y,\mathbf z) =&  || \mathbf{b}-\mathbf{y}||_1 + i_{\boldsymbol{\mathcal{S}}}(\mathbf{x}) \notag \\
&+ \langle\mathbf{z},\mathbf{y} - \mathbf{A}\mathbf{x}\rangle + \frac{\beta}{2} \|\mathbf{y} - \mathbf A \mathbf x \|^2_2,
\end{align}
where $\beta > 0$ is a penalty parameter.
Based on ADMM, update rules are given by
\begin{align}
  \mathbf x &\leftarrow \mathop{\text{argmin}}_{\mathbf{x}} L_{\rm L1}(\mathbf x,\mathbf y,\mathbf z), \\
  \mathbf y &\leftarrow \mathop{\text{argmin}}_{\mathbf{y}} L_{\rm L1}(\mathbf x,\mathbf y,\mathbf z), \\
  \mathbf z &\leftarrow \mathbf z + \beta(\mathbf y - \mathbf A \mathbf x).
\end{align}

Sub-optimization for updating $\mathbf x$ can be reduced to follow:
\begin{align}
  &\mathop{\text{argmin}}_{\mathbf{x}} \ -\langle\mathbf{z},\mathbf{A}\mathbf{x}\rangle + \frac{\beta}{2} \|\mathbf{y} - \mathbf A \mathbf x \|^2_2 + i_{\boldsymbol{\mathcal{S}}}(\mathbf{x}) \notag \\
  &= \mathop{\text{argmin}}_{\mathbf{x}}  \left\| \left(\mathbf{y} + \frac{1}{\beta}\mathbf{z}\right) - \mathbf{A}\mathbf{x} \right\|^2_2 + i_{\boldsymbol{\mathcal{S}}}(\mathbf{x}). \label{eq:ADMM_update_x}
\end{align}
Since \eqref{eq:ADMM_update_x} is same form to \eqref{eq:MMobj}, it can be solved by using MM iterations.
This MM iteration can be given by replacing $\mathbf b$ in \eqref{eq:L2MMupdate} with $\mathbf y + \frac{1}{\beta} \mathbf z$.
In the sense of argmin, it requires many iterations of MM update, but we just substitute only one MM update in ADMM.

Sub-optimization for $\mathbf y$ has closed-form solution:
\begin{align}
  &\mathop{\text{argmin}}_{\mathbf{y}} || \mathbf b - \mathbf y ||_1 + \langle\mathbf{z},\mathbf{y}\rangle + \frac{\beta}{2} \|\mathbf{y} - \mathbf A \mathbf x \|^2_2  \notag \\
  &= \mathop{\text{argmin}}_{\mathbf{y}} \frac{1}{\beta}|| \mathbf b - \mathbf y ||_1 +  \frac{1}{2} \left\| \left(\mathbf A\mathbf{x} - \frac{1}{\beta}\mathbf{z}\right) - \mathbf{y} \right\|^2_2 \notag \\
  &= \mathbf b + \mathcal{T}_{\frac{1}{\beta}}\left[\mathbf A\mathbf{x} - \frac{1}{\beta}\mathbf{z} - \mathbf b \right],
  \label{eq:L1_ADMM_update_y}
\end{align}
where $\mathcal{T}_\rho[\mathbf v](i) := \text{sign}(v_i) \max( | v_i | - \rho, 0)$ is an entry-wise soft-thresholding operator with threshold $\rho > 0$.

\subsubsection{ADMM for KL divergence}\label{sec:KLcase}
For two non-negative variables $\mathbf b \in \mathbb{R}^{I}_+$ and $\mathbf y \in \mathbb{R}^I_+$, KL divergence (I-divergence) is defined by
\begin{align}
  D_{\rm{KL}}(\mathbf{b},\mathbf{y}) := \sum_{i=1}^I b_i \log \frac{b_i}{y_i} + y_i - b_i.
\end{align}

Employing KL divergence for \eqref{eq:general_tensor_reconstruction_problem}, the problem is given as
\begin{align}
\mathop{\text{minimize}}_{\mathbf x} D_{\rm{KL}}(\mathbf{b},\mathbf A\mathbf{x}) + i_+(\mathbf A\mathbf x)+ i_{\boldsymbol{\mathcal{S}}}(\mathbf x),
\end{align}
where $i_+(\cdot)$ is an indicator function for non-negative constraint $\mathbf A\mathbf x \in \mathbb{R}^{I}_+$.
It can be transformed into the following equivalent problem:
\begin{align}
\mathop{\text{minimize}}_{\mathbf x, \mathbf y} D_{\rm{KL}}(\mathbf{b},\mathbf{y}) + i_+(\mathbf y)+ i_{\boldsymbol{\mathcal{S}}}(\mathbf x), \text{s.t. } \mathbf y = \mathbf A \mathbf x.
\end{align}
Its augmented Lagrangian is given by
\begin{align}
L_{\rm KL}(\mathbf x,\mathbf y,\mathbf z) =&  D_{\rm{KL}}(\mathbf{b},\mathbf{y}) + i_+(\mathbf y) + i_{\boldsymbol{\mathcal{S}}}(\mathbf{x}) \notag \\
&+ \langle\mathbf{z},\mathbf{y} - \mathbf{A}\mathbf{x}\rangle + \frac{\beta}{2} \|\mathbf{y} - \mathbf A \mathbf x \|^2_2,
\end{align}
Optimization steps of ADMM with KL divergence can be derived in similar way to $\ell_1$ case, just replacing $\ell_1$-loss with KL divergence plus indicator function.
Update rule for $\mathbf x$ is the same as $\ell_1$-case since augmented Lagrangian is completely equivalent with respect to $\mathbf x$.
Only the update rule for $\mathbf y$ is different.

Sub-optimization for updating $\mathbf y$ is given as
\begin{align}
  \mathop{\text{argmin}}_{\mathbf{y} > \mathbf 0} \frac{1}{\beta}D_{\rm{KL}}(\mathbf{b},\mathbf{y}) +  \frac{1}{2} \left\| \left(\mathbf A\mathbf{x} - \frac{1}{\beta}\mathbf{z}\right) - \mathbf{y} \right\|^2_2.
  \label{eq:KL_ADMM_update_y}
\end{align}
Note that the objective function is separable to each $y_i$, and its closed-form solution can be given by
\begin{align}
  \hat{y}_i = \frac{1}{2\beta}(\beta d_i - 1 + \sqrt{ (\beta d_i -1)^2 + 4\beta b_i }),
\end{align}
where we put $\mathbf d = \mathbf A \mathbf x - \frac{1}{\beta} \mathbf z$.

\begin{algorithm}[t]
\caption{ADMM-MM algorithm} \label{alg2}
\begin{algorithmic}[1]  
    \STATE \textbf{input}: $\mathbf b$, $\mathbf A$, $\boldsymbol{\mathcal S}$, type of loss, $\alpha$, $\beta > 0$
    \STATE Initialize $\mathbf x \in \boldsymbol{\mathcal S}$ and $\mathbf z = \mathbf 0$;
    \STATE $\lambda \leftarrow$ maximum eigenvalue of $\mathbf{A}^\top\mathbf{A}$;
    \STATE
    $
\gamma \leftarrow \begin{cases} 
    \frac{\alpha}{\lambda} &(\text{$\ell_2$})\\
    \frac{2\alpha}{\beta \lambda} &(\text{$\ell_1$ or KL})\\
    \end{cases} 
$
     \REPEAT
     \STATE $\mathbf{d} \leftarrow \mathbf{A}\mathbf{x}-\frac{1}{\beta}\mathbf{z}$; 
\STATE 
$
y_i \leftarrow \begin{cases} 
    b_i &(\text{$\ell_2$})\\
    b_i + \mathcal{T}_{\frac{1}{\beta}}(d_i - b_i) &(\text{$\ell_1$})\\
    \frac{\beta d_i - 1 + \sqrt{(\beta d_i - 1)^2 + 4\beta b_i}}{2\beta}&(\text{KL})\\
    \end{cases} 
$

\STATE 
$
\mathbf{z} \leftarrow \begin{cases}
  \mathbf 0 &(\text{$\ell_2$})\\
\mathbf{z} + \beta(\mathbf{y} - \mathbf{A}\mathbf{x})&(\text{$\ell_1$ or KL})\\
    \end{cases}
$

\STATE
$
\mathbf{x} \leftarrow \rm{proj}_{\boldsymbol{\mathcal{S}}_\gamma}\left[\mathbf{x} - \frac{1}{\lambda}(\mathbf{A}^\top\mathbf{A}\mathbf{x} - \mathbf{A}^\top(\mathbf{y} +  \frac{1}{\beta}\mathbf{z}))\right]
$;
     \UNTIL{convergence}
     \STATE \textbf{output}: $\mathbf{x}$ 
\end{algorithmic}
\end{algorithm}

\subsection{Penalized TD}
In some cases, we penalize parameters with TD such as for sparse, smooth, and Tikhonov regularization.
Let $p(\theta)$ and $\alpha$ be a penalty function and its trade-off parameter, the LS-based TD problem is modified to
\begin{align}
\hat{\theta}_\alpha = \mathop{\text{argmin}}_{\theta \in \Theta} || \text{vec}(\boldsymbol{\mathcal{V}}) - \text{vec}(\boldsymbol{\mathcal{X}}(\theta)) ||_2^2 + \alpha p(\theta).
\end{align}
In this case, a mapping from $\text{vec}(\boldsymbol{\mathcal{V}})$ to $\text{vec}(\boldsymbol{\mathcal{X}}(\hat{\theta}_\alpha))$ is clearly different from \eqref{eq:LSTD} with $\alpha  \neq 0$, but we also consider it as a {\it projection} onto some set $\boldsymbol{\mathcal{S}}_\alpha$.
We denote it as follow:
\begin{align}
  \text{vec}(\boldsymbol{\mathcal{X}}(\hat{\theta}_\alpha)) = \mathop{\text{argmin}}_{\mathbf x \in \boldsymbol{\mathcal{S}}} || \mathbf v - \mathbf x ||_2^2 + \alpha p(\theta) =: \text{proj}_{\boldsymbol{\mathcal{S}}_\alpha}(\mathbf v). \label{eq:LSTDp}
\end{align}
Note that \eqref{eq:LSTDp} includes \eqref{eq:LSTD} as a case of $\alpha = 0$.

In the case of penalized TD, MM iterations for $\ell_2$-loss is given by
\begin{align}
  \mathbf x_{t+1} = \text{proj}_{\boldsymbol{\mathcal{S}}_{\alpha/\lambda}}\left[ \mathbf{x}_t - \frac{1}{\lambda}(\mathbf{A}^\top\mathbf{A}\mathbf{x}_t - \mathbf{A}^\top \mathbf{b}) \right]. \label{eq:L2MMupdate_p}
\end{align}
It can be easily derived from \eqref{eq:aux_func} with additional term $\alpha p(\theta)$.

Furthermore, sub-optimization problem for updating $\mathbf x$ in ADMM can be given by 
\begin{align}
\mathop{\text{argmin}}_{\mathbf{x}}  \left\| \left(\mathbf{y} + \frac{1}{\beta}\mathbf{z}\right) - \mathbf{A}\mathbf{x} \right\|^2_2 + i_{\boldsymbol{\mathcal{S}}}(\mathbf{x}) + \frac{2\alpha}{\beta} p(\theta). \label{eq:ADMM_update_x_p}
\end{align}
This is \eqref{eq:ADMM_update_x} with a penalty term.
Its MM iteration is given by
\begin{align}
\mathbf{x}_{t+1} \leftarrow \rm{proj}_{\boldsymbol{\mathcal{S}}_{\frac{2\alpha}{\beta\lambda}}}\left[\mathbf{x}_t - \frac{1}{\lambda}(\mathbf{A}^\top\mathbf{A}\mathbf{x}_t - \mathbf{A}^\top(\mathbf{y} +  \frac{1}{\beta}\mathbf{z}))\right].
\end{align}

\subsection{Unified Algorithm}
Finally, the proposed method can be summarized in Algorithm~\ref{alg2}.
We only need to change update formulas for $\mathbf y$ and $\mathbf z$ to match the type of loss function.
At the 8th line, the projection onto $\boldsymbol{\mathcal{S}}$ can be replaced by any TD algorithm in a kind of plug-and-play manner.
However, note that it is inefficient to perform a full TD in each iteration, and the projection onto $\boldsymbol{\mathcal{S}}$ is usually replaced by one iteration of ALS.
Although it is not written explicitly, it is necessary to retain the core tensors and factor matrices, and use and update them in each iteration.

\subsection{Convergence}
In general, tensor decompositions are non-convex problem and there is no guarantee of global convergence.
There are only several algorithms having properties of local convergence such as ALS, hierarchical ALS (HALS) \cite{cichocki2007nmf}, multiplicative update (MU) \cite{lee2000algorithms} and BCD with sufficiently small step-size.

In our study, the algorithm with $\ell_2$-loss using ALS, HALS, or MU has local convergence.
Each ALS/HALS/MU update monotonically decreases (non-increases) auxiliary function and it achieves monotonic decreasing (non-increasing) property of original objective function by MM framework \cite{sun2016majorization}.
ALS is a workhorse algorithm used for many LS-based TDs (CP, Tucker, TT, TR), and HALS is often employed for LS-based CP decomposition and non-negative matrix/tensor factorization (NMF/NTF) \cite{kolda2009tensor,cichocki2007nmf}.
Most of MU rules for NMF/NTF are derived by MM framework \cite{lee2000algorithms,cichocki2007nmf,gillis2020nonnegative}.

In cases of $\ell_1$-loss and KL-loss, we have no result of convergence.
Usually the convergence of ADMM is based on the non-expansive property of projection onto convex set or proximal mapping of convex function \cite{eckstein1992douglas}.
In the proposed algorithm, LS-based TD is a projection onto non-convex set and it does not match above condition.

\begin{table}[t]
\centering
\caption{Comparison of objective functions and time [sec] at convergence for PG, BCD, and Proposed methods}
\scriptsize
\begin{tabular}{|cc|cc|cc|cc|}
\hline
\multicolumn{2}{|c|}{} &\multicolumn{2}{c|}{PG}&\multicolumn{2}{c|}{BCD}&\multicolumn{2}{c|}{Proposed}\\ 
\multicolumn{2}{|c|}{}& obj. & time & obj. & time & obj. & time\\  \hline
\multirow{4}{*}{$\ell_2$}
& \multicolumn{1}{|c|}{noise} & - & - & 1623 & 410.3 & \textbf{1614} & \textbf{8.924}\\  
& \multicolumn{1}{|c|}{missing} & - & - & 8091 & 1866 & \textbf{119.0} & \textbf{7.792} \\ 
& \multicolumn{1}{|c|}{blur}& - & - & 4657 & 13684 & \textbf{26.10} & \textbf{298.5}\\ 
& \multicolumn{1}{|c|}{down}& - & - & 80.68 & 1165 & \textbf{2.365} &\textbf{11.36} \\  \hline
\multirow{4}{*}{$\ell_1$}
& \multicolumn{1}{|c|}{noise} & 42.92 & 13.02 & 43.18 & 41.66 & \textbf{42.87} & \textbf{2.16} \\ 
& \multicolumn{1}{|c|}{missing}& 39.74 & \textbf{48.92} & 38.65 & 89.79 & \textbf{35.09} & 84.9\\ 
& \multicolumn{1}{|c|}{blur}& 39.14 & \textbf{103.9} & 39.57 & 171.7 & \textbf{38.93} & 133.8\\ 
& \multicolumn{1}{|c|}{down}& 1.040 & 108.9 & 1.759 & 42.18 & \textbf{1.004}  & \textbf{4.85}\\  \hline
\multirow{4}{*}{KL}
& \multicolumn{1}{|c|}{noise} & 5.361 & 20.50 & 5.394 & 146 & \textbf{5.358}& \textbf{4.65}\\ 
& \multicolumn{1}{|c|}{missing}& 0.526 & 54.95 & 0.363 & 220.8 & \textbf{0.247} & \textbf{8.532}\\ 
& \multicolumn{1}{|c|}{blur}& 5.109 & \textbf{158.2} & 5.155 & 576.4 & \textbf{4.905} & 669.3\\  
& \multicolumn{1}{|c|}{down}& 0.189 & 98.59 & 0.304 & 32.16 & \textbf{0.135} & \textbf{7.65}\\ \hline
\end{tabular} 
\label{tb:cost_time}
\end{table}

\begin{figure*}[t]
\centering
\begin{minipage}{.325\textwidth}
\centering
\includegraphics[width = \textwidth]{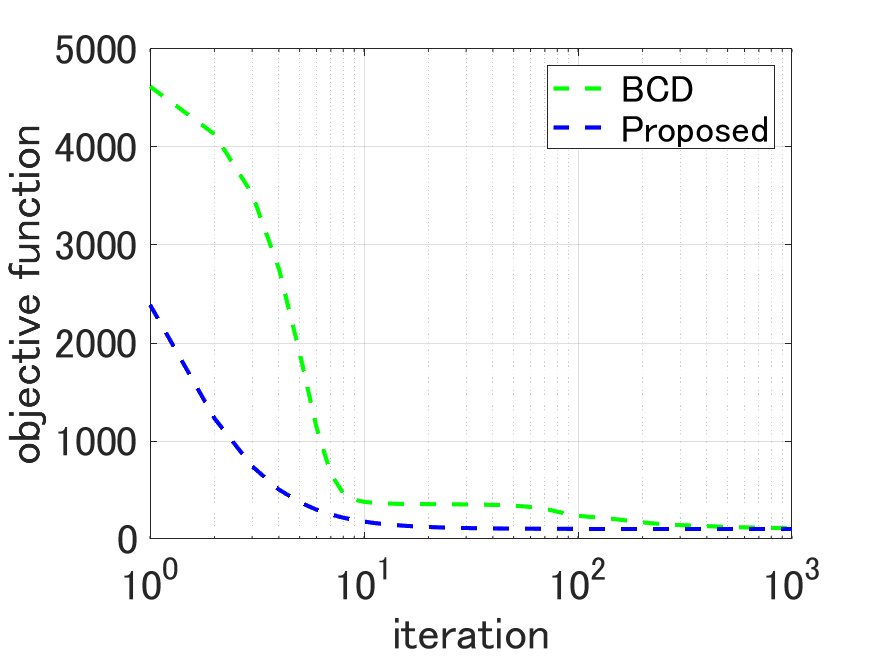}
(a) $\ell_2$-loss
\end{minipage}
\begin{minipage}{.325\textwidth}
\centering
\includegraphics[width =\textwidth]{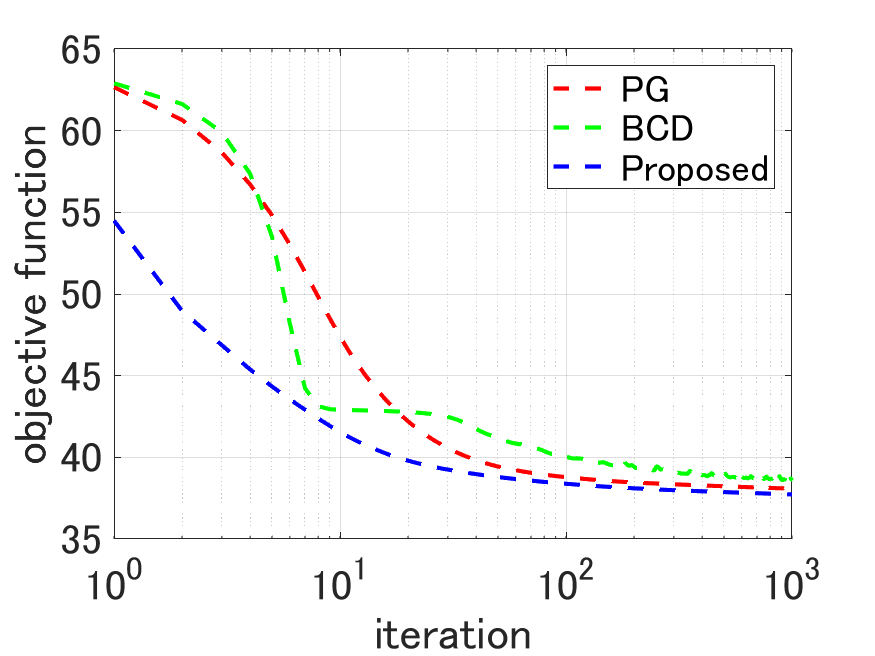}
(b) $\ell_1$-loss
\end{minipage} 
\begin{minipage}{.325\textwidth}
\centering
\includegraphics[width = \textwidth]{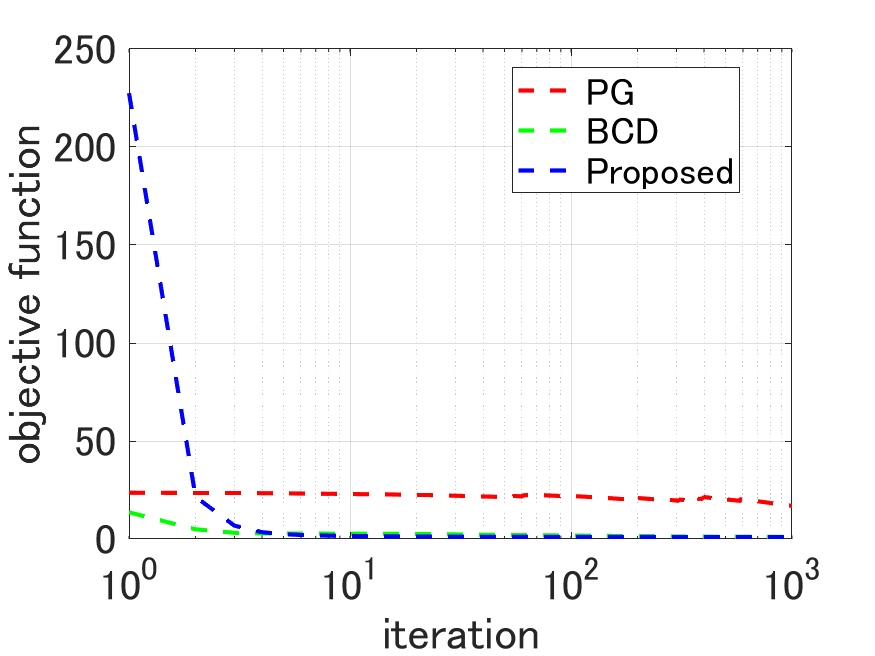}
(c) KL divergence
\end{minipage}
\caption{Optimization behaviors: (a), (b), and (c) show comparison of PG, BCD and the proposed algorithms for various loss functions in CP based tensor completion task.}\label{fig:opt_behav}
\end{figure*}

\begin{figure}
\centering
\includegraphics[width = 0.35\textwidth]{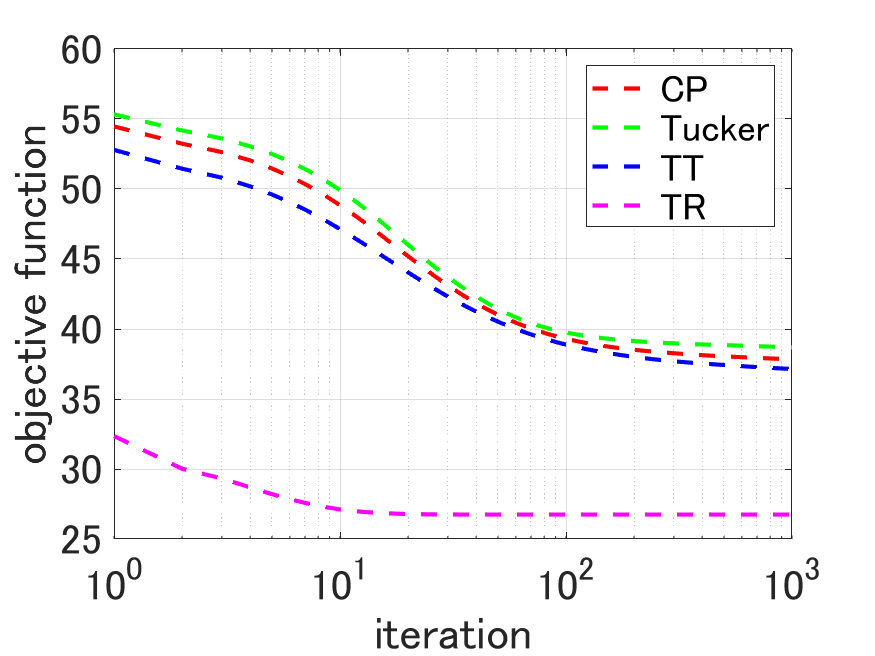}
\caption{Application for various TD models}\label{fig:opt_behav_TDs}
\end{figure}

\section{Related Works}
There are several studies of algorithms for TDs using ADMM.
AO-ADMM \cite{huang2016flexible} is an algorithm for constrained matrix/tensor factorization which solves the sub-problems for updating factor matrices using ADMM.
Although alternating optimization (AO) is the main-routine and ADMM is subroutine in AO-ADMM, in contrast, the proposed ADMM-MM algorithm is ADMM is used as main-routine.
AO-ADMM support several loss function, but it does not support various design matrices.
In addition, AO-PDS \cite{ono2018efficient} has been proposed using primal-dual splitting (PDS) instead of ADMM.

In \cite{zhang2013robust}, robust Tucker decomposition (RTKD) with $\ell_1$-loss has been proposed and its algorithm has been developed based on ADMM.
RTKD employs ADMM as main-routine and ALS is used as subroutine, and it can be regarded as a special case of our proposed algorithm.
RTKD does not support any other loss function, various design matrices, and other constraints.

ADMM-based NMF algorithms \cite{xu2012alternating,sun2014alternating,hajinezhad2016nonnegative} have been studied.
These algorithms are slightly different from AO-ADMM \cite{huang2016flexible} and our ADMM-MM because they incorporate alternating updates of the factor matrices in the same sequence in the ADMM iterations.
The algorithm is unstructured, it is difficult to generalize and extend.

In the context of generalized tensor decomposition, generalized CP tensor decomposition \cite{hong2020generalized} has been proposed. The purpose of \cite{hong2020generalized} is to make CP decomposition compatible with various loss functions. Basically, a BCD-based algorithm has been proposed. However, other TD models and perspectives on design matrices are not discussed.

Plug-and-play (PnP)-ADMM \cite{venkatakrishnan2013plug,chan2016plug} is a framework for using some black-box models (e.g., trained deep denoiser) instead of proximal mapping in ADMM.
It is highly extensible in that any model can be applied to various design matrices.
The structure of using LS-based TD in a plug-and-play manner in the proposed algorithm is basically the same as PnP-ADMM.
If we consider tensor decomposition as a denoiser, the proposed algorithm may be considered a type of PnP-ADMM.
In this sense, the proposed algorithm and PnP-ADMM are very similar, but they are significantly different in that our objective function is not black box.

\section{EXPERIMENT}

\subsection{Data and Comparison Methods}
In this experiment, an RGB image of size $256\times256\times3$ is used as a ground-truth low-rank 3rd order tensor $\mathbf x_0$.
Observation signals $\mathbf b$ are artificially generated using various design matrices $\mathbf A$ and noise $\mathbf n$ according to the equation $\mathbf b = \mathbf A \mathbf x_0 + \mathbf n$.

We compare the proposed algorithm with PG and BCD algorithms for CP based GTD.
PG iteration is given by
\begin{align}
  \mathbf x \leftarrow \text{proj}_{\boldsymbol{\mathcal{S}}}\left[\mathbf x - \mu \nabla_{\mathbf x} D(\mathbf b, \mathbf A\mathbf x) \right];
\end{align}
and BCD iteration is given by
\begin{align}
  &\mathbf U_1 \leftarrow \mathbf U_1 - \mu \nabla_{\mathbf U_1} D(\mathbf b, \mathbf A \mathbf x(\mathbf U_1, \mathbf U_2, \mathbf U_3));\\
  &\mathbf U_2 \leftarrow \mathbf U_2 - \mu \nabla_{\mathbf U_2} D(\mathbf b, \mathbf A \mathbf x(\mathbf U_1, \mathbf U_2, \mathbf U_3));\\
  &\mathbf U_3 \leftarrow \mathbf U_3 - \mu \nabla_{\mathbf U_3} D(\mathbf b, \mathbf A \mathbf x(\mathbf U_1, \mathbf U_2, \mathbf U_3));
\end{align}
where $\mu > 0$ is a step-size.
The gradients are computed by using auto-gradient function in MATLAB, and $\mu$ was manually adjusted for the best performance.

\subsection{Optimization behavior}
We first compare optimization behaviors of the proposed ADMM-MM algorithm with PG and BCD for general CP decomposition.
We used four types of design matrices and three types of loss functions in this experiments.
Table~\ref{tb:cost_time} shows the achieved values of objective function and its computational time [sec] for the three optimization methods to various settings.
The best values are highlighted in bold.
The proposed method reduces the objective function stably and efficiently in various settings than PG and BCD.
Fig.~\ref{fig:opt_behav}(a)-(c) show its selected optimization behaviors on the missing task based on three loss functions.
Note that, since the proposed method for $\ell_2$-loss and PG are equivalent, they are not compared.

Fig.~\ref{fig:opt_behav_TDs} shows the optimization behavior of the proposed method for $\ell_1$-loss with missing design matrix when applying various TD models: CP, Tucker, TT, and TR.
All LS-based TDs can be optimized by ALS, and we used one cycle of ALS in the proposed ADMM-MM algorithm.
It can be seen that the proposed algorithm works well for minimizing $\ell_1$-loss with various TD models.

\begin{figure*}[t]
\centering
\includegraphics[width=0.8\textwidth]{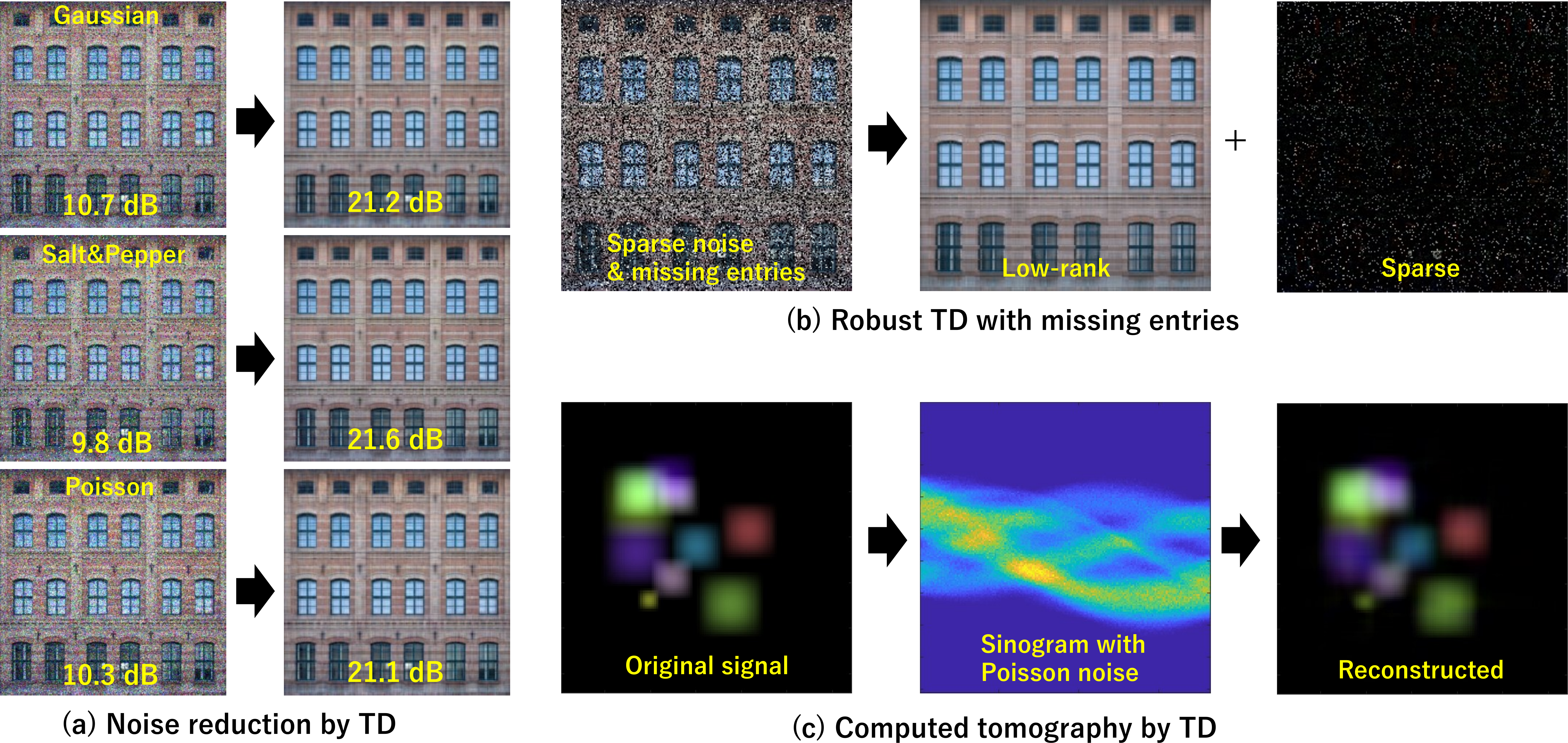}
\caption{Application to image reconstruction by using proposed optimization with SPC model \cite{yokota2016smooth}. }
 \label{fig:proposed_Images}
\end{figure*}

\subsection{Image Processing Applications}
We applied the proposed algorithm to the smooth PARAFAC (SPC) model \cite{yokota2016smooth}.
Although SPC model has been proposed originally for LS-based tensor completion, our framework can extend SPC model into $\ell_1$-loss and KL-divergence with arbitrary design matrix $\mathbf A$.
Since SPC model employs HALS algorithm for its optimization, we use one cycle of HALS updates for LS-based TD module (9th line of Algorithm~\ref{alg2}) in ADMM-MM algorithm.

Fig.~\ref{fig:proposed_Images} shows the results of image processing tasks: noise reduction for various distributions, robust TD with missing entries, and computed tomography.
Noise reduction tasks with various noise distributions can be provided by setting $\mathbf A = \mathbf I$ and selecting appropriate loss functions.
Robust TD with missing entries can be provided by setting $\mathbf A$ as an entry elimination matrix and select $\ell_1$-loss.
Computed tomography can be provided by setting $\mathbf A$ as Radon transform matrix and select KL-loss.


\section{CONCLUSION}
In this study, we proposed a new optimization algorithm for GTD.
The proposed GTD algorithm supports three loss functions with arbitrary design matrix, and any LS-based TD algorithm
can be easily extended into GTD problem setting in a plug-and-play manner.
This framework can provide a situation where we can focus only on the LS-based TD.
Many TD models have been studied by LS setting at first, and their algorithms are often efficient such as ALS.
We hope that the contribution of this study will not be limited to this paper, but will lead to a wide range of applications in the future.

\bibliographystyle{IEEE}

\end{document}